\documentclass{article}
\usepackage{spconf,amsmath,epsfig}
\usepackage{booktabs}
\usepackage{makecell}
\usepackage{stfloats}
\usepackage{color}
\usepackage{subcaption}

\newcounter{daggerfootnote}

\title{FINE-GRAINED BUILDING ROOF INSTANCE SEGMENTATION BASED ON DOMAIN ADAPTED PRETRAINING AND COMPOSITE DUAL-BACKBONE}
\name{\parbox{\linewidth}{\centering Guozhang Liu$^{\dagger}$\thanks{ $^{\dagger}$ Equal contributions}, Baochai Peng$^{\dagger}$, Ting Liu, Pan Zhang, Mengke Yuan, \\ Chaoran Lu, Ningning Cao, Sen Zhang, Simin Huang, Tao Wang$^{*}$\thanks{ $^{*}$ Corresponding author, wangtao@piesat.cn}}}

\address{PIESAT Information Technology Co, Ltd., Beijing, China}

\begin{document}
\maketitle

\begin{abstract}
The diversity of building architecture styles of global cities situated on various landforms, the degraded optical imagery affected by clouds and shadows, and the significant inter-class imbalance of roof types pose challenges for designing a robust and accurate building roof instance segmentor. To address these issues, we propose an effective framework to fulfill semantic interpretation of individual buildings with high-resolution optical satellite imagery. Specifically, the leveraged domain adapted pretraining strategy and composite dual-backbone greatly facilitates the discriminative feature learning. Moreover, new data augmentation pipeline, stochastic weight averaging (SWA) training and instance segmentation based model ensemble in testing are utilized to acquire additional performance boost. Experiment results show that our approach ranks in the first place of the 2023 IEEE GRSS Data Fusion Contest (DFC) Track 1 test phase ($mAP_{50}$:50.6\%). Note-worthily, we have also explored the potential of multimodal data fusion with both optical satellite imagery and SAR data. 
\end{abstract}

\section{introduction}
Automated and in-depth building interpretation is meaningful in remote sensing scenario, as the investigation results are of great significance in urban planning, smart city, and emergency management applications~\cite{huang2022urban}. Unfortunately, the growing data volume and rapidly evolving deep learning techniques can still hardly meet the requirements of practical building roof detection and fine-grained roof type classification due to the lack of finely-annotated and well-organized dataset, the variable instance sizes, the diverse architectural styles and the great inter-class quantity discrepancy, etc.

\begin{figure}[t]
\centering
\includegraphics[width=1\linewidth]{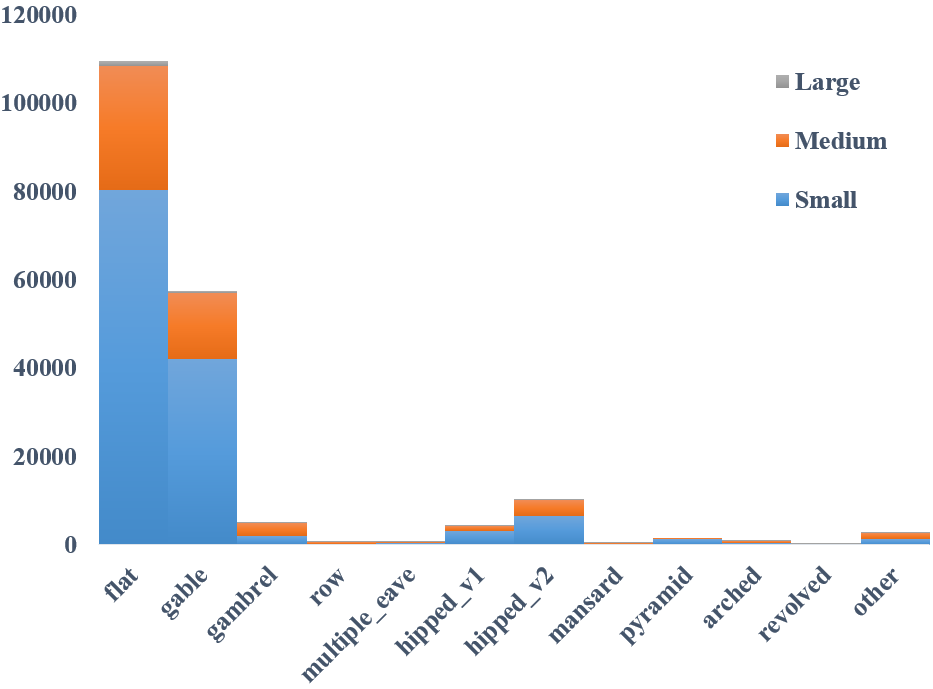}
\caption{Distribution of instance pixels among predefined roof categories of DFC 2023. Gray, orange and blue represent large, medium and small instances respectively.}
\label{fig:statistic}
\end{figure}

\begin{figure}[t]
\centering
\includegraphics[width=1\linewidth]{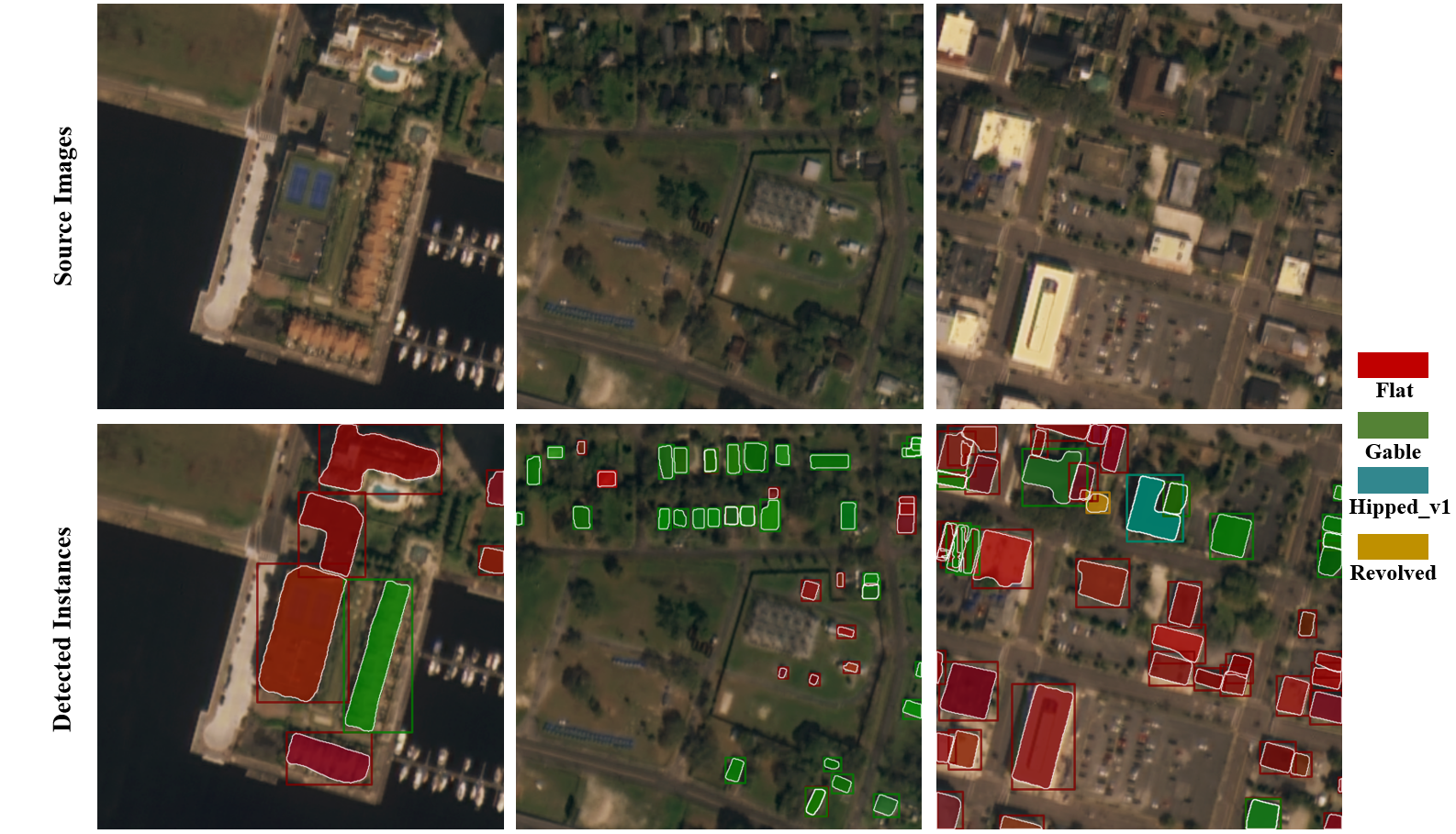}
\caption{This figure displays our inference results (the second row) for satellite imagery input (the first row).}
\label{fig:results}
\end{figure}

The DFC 2023~\cite{Persello2023} has constructed a large-scale, multi-modal and elaborately labeled dataset in promoting the performance of building detection and roof type classification methods. The dataset contains a total of 12 predefined roof categories and about 190,000 high-quality mask annotations. The statistics of track1 training dataset (Fig.~\ref{fig:statistic}) reflects practical difficulties that there are significant proportion disparities of different classes and high-density small targets. The instances of the largest category ``Flat roof'' are more than 200 times of the least category ``Revolved roof''. Small objects account for 71$\%$ of the total number of objects. Huang et al.~\cite{huang2022urban} validate the representative single-stage (SOLOv2), two-stage (Mask-RCNN, Cascade Mask RCNN), and query-based (QueryInst) methods which can not achieve desirable performance in similar UBC~\cite{huang2022urban} dataset. In summary, the three challenges in fine-grained roof instance segmentation are: \textbf{1) the long-tail distribution of diverse roof styles}; \textbf{2) the detection of crowded small objects}; \textbf{3) the ambiguous visual features among different categories}. 

This paper proposes a simple but robust instance segmentation framework based on Cascade Mask R-CNN~\cite{cai2018cascade} with domain-adapted pretraining and modernized dual-backbone to address these problems. We first use the domain adaptive pretraining model to initialize the parameters to improve the stability. Secondly, we adopt a composite dual-branch backbone structure to construct a more robust and discriminative feature extractor, which alleviates the improper segmentation for small-size instances and misclassification of minority categories. The dual-branch backbone can also facilitate multi-modal data fusion. Moreover, dedicated data augmentation pipeline with modified copy-paste, SWA\cite{zhang2020swa} training strategy, and an instance segmentation model aggregation in inference jointly improve the precision of our fine-grained roof instance segmentor. Our visual instance segmentation results are displayed in Fig.~\ref{fig:results}, in which small and rare (Revolved) objects can both be detected accurately. Comprehensive experiments and ablation study demonstrate the superiority of proposed method, which has achieved a $mAP_{50}$ of 50.6\% and won the first place on track 1 test phase of DFC 2023. 

\begin{figure*}[ht]
\centering
\includegraphics[width=0.9\linewidth]{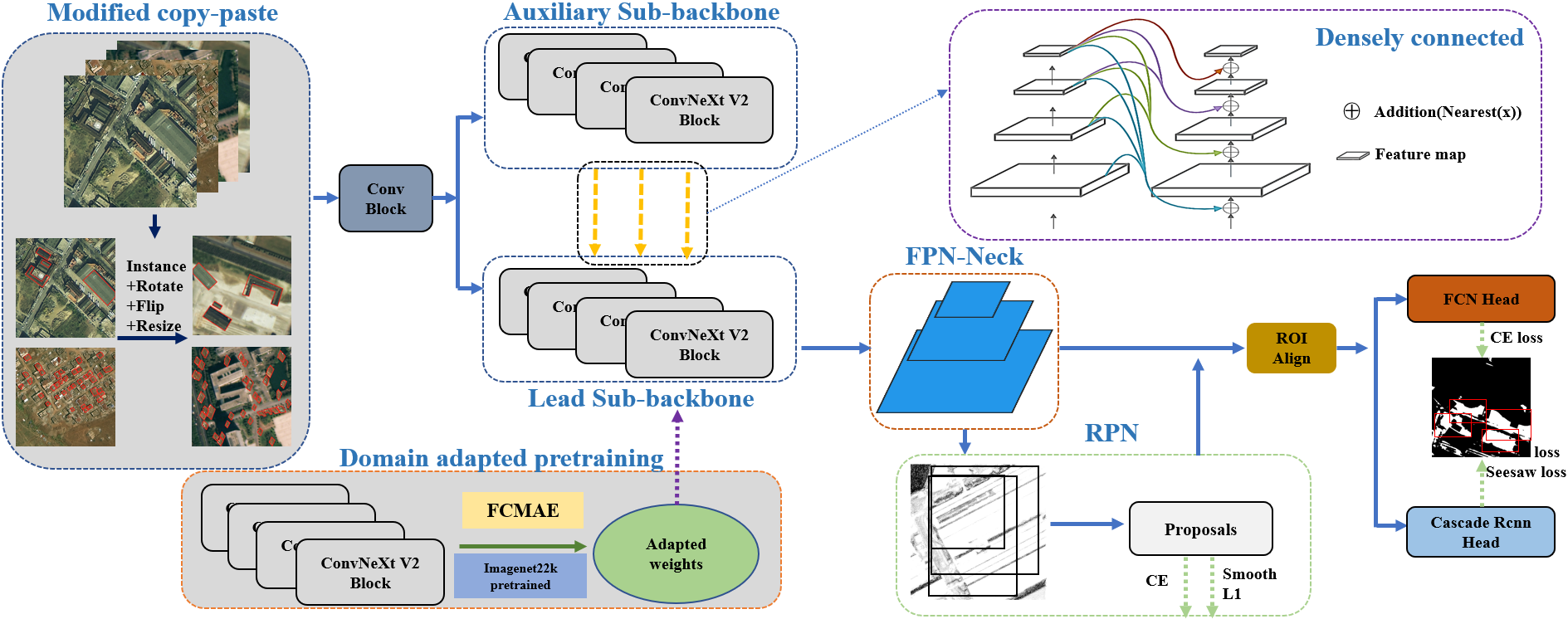}
\caption{The overview training workflow and network structure diagram of our proposed roof instance segmentor.}
\label{fig:framwork}
\end{figure*}

\section{Methods}
\label{sec:methods}

This section will detail the proposed fine-grained roof instance segmentation framework. The overall network structure diagram is shown in Fig.~\ref{fig:framwork} armed with ConvNeXt V2~\cite{Woo2023ConvNeXtV2} based composite dual-backbone and domain adapted pretraining, modified copy-paste data augmentation.  

\subsection{Composite dual-backbone feature extractor}
\label{sssec:subhead}
For the purpose of constructing a robust feature extractor, we adopt a dual-branch structure backbone containing two densely connected sub-backbones, which is validated in CBNet~\cite{liang2022cbnetv2}. Intuitively, it is a flexible structure for both single-modality and muti-modality input. Feature maps from all higher stages of the auxiliary sub-backbone are nearest interpolated and added to lower-level stages of the lead sub-backbone, as illustrated in equation (1). This densely connected structure deeply fuses the high-level semantic information and low-level semantic information of two sub-backbones to boost our feature extractor. Furthermore, there are some competitive alternativs to our sub-backbone, such as ConvNeXt V2 \cite{Woo2023ConvNeXtV2}, Swin Transformer \cite{liu2021swin}.
\begin{equation}
F_{lead}^{j} =G_{lead}^{j}(\sum_{i=j}^{L}N(F_{aux}^{i}) + {F}_{lead}^{j-1})
\label{equ:backbone}
\end{equation}
Where $L$ represents the total stage number, $F_{aux}^{i}$, $F_{lead}^{j}$ represent the feature map of $i_{th}$, $j_{th}$ stage in the auxiliary sub-backbone and lead sub-backbone, $G_{lead}^{j}(x)$ represents model block of $j_{th}$ stage in lead the sub-backbone, N represents nearest interpolation function.

\subsection{Domain adapted pretraining}
\label{sssec:subhead1}
Model weights initialization strategy plays a key role in the entire optimization. Typically, Imagenet22k pretrained weights can not only accelerate the convergence, but also improve the performance. In the training process, we attempt to employ a better initialization through domain adaptation pretraining. Fully convolutional masked autoencoder (FCMAE)~\cite{Woo2023ConvNeXtV2} is a self-supervised pretraining method for conv-based models, which is beneficial for the capability of domain adaption on specific dataset. Therefore, we pretrain the ConvNeXt V2 weights on the RGB modality dataset and then initialize our two sub-backbones with pretraining weights.

\subsection{Modified copy-paste}
\label{sssec:subhead2}

\begin{figure}[htb]
\centering
\begin{minipage}[t]{0.3\linewidth}
  \centering
  \centerline{\epsfig{figure=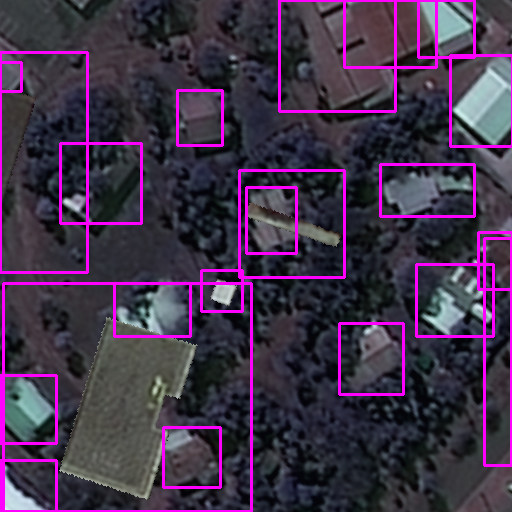,width=2.3cm}}
  \vspace{0cm}
\end{minipage}
\begin{minipage}[t]{0.3\linewidth}
  \centering
  \centerline{\epsfig{figure=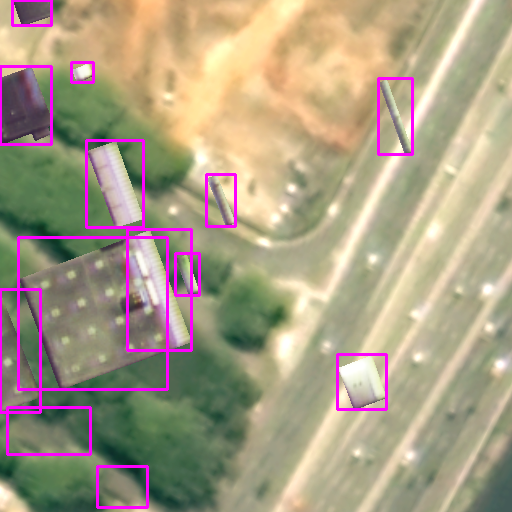,width=2.3cm}}
  \vspace{0cm}
\end{minipage}
\begin{minipage}[t]{0.3\linewidth}
  \centering
  \centerline{\epsfig{figure=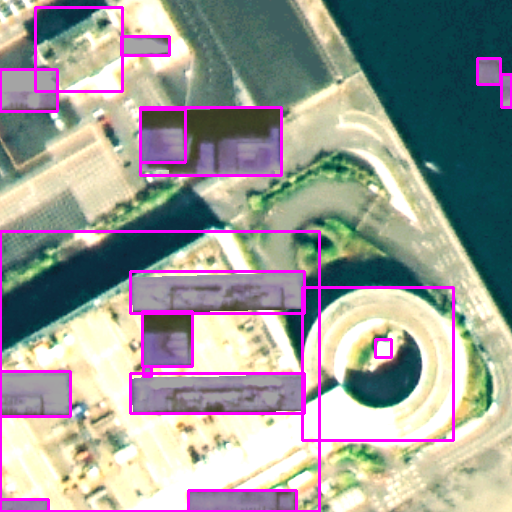,width=2.3cm}}
  \vspace{0cm}
\end{minipage}

\caption{Synthetic images produced by data augmentation.}
\label{fig:cpoy_paste}
\end{figure}

Simple copy-paste \cite{zhang2020simple} is an effective way of instance-level data augmentation. To further enrich the dataset, we propose a modified version which first crops instances from images, and then apply random resize, rotation, and flip to these pixel-level instances before pasting them onto the chosen images. These processes can hardly damage the semantic information in the nadir-viewing images. Synthetic images can be found in Fig.~\ref{fig:cpoy_paste}. To prevent overall distribution shift caused by synthetic images, we separate the training process into two phases: 1) training with modified copy-paste data for the first 90\% total epochs; 2) fine-tuning without the modified copy-paste data for the last 10\% total epochs. In addition our data augmentation pipeline contains some regular data augmentation strategies like random rotate, random resize, random crop, etc.

\subsection{Model ensemble}
\label{sssec:subhead3}
To further enhance the performance of our model, we employ a model ensemble strategy to fuse several different models. For object detection, weighted bounding boxes fusion (WBF)~\cite{solovyev2021weighted} is a popular approach for model ensemble. We modify WBF as Weightd Segmentation Fusion(WSF), which is suitable for instance segmentation. First, we adopt the WBF strategy to obtain fused bounding boxes, then fuse masks from different models to get the final results.

\subsection{Additional training strategy}
\label{sssec:subhead4}
In order to address the issues of long-tail distribution of roof categories and distribution disparity between training and validation datasets, we try to build a more robust model through various training techniques.

\textbf{Tactics for long-tail distribution:} We adopt a balanced sample strategy to increase the probability of samples containing tail classes of being chosen. Additionally, we use seesaw loss~\cite{seesawloss} to reduce the gradients from negative samples on tail classes.

\textbf{Stochastic Weight Average:} We employ the SWA technique with cyclical learning rates to train for the next 12 epochs after all training epochs. Subsequently, we average these 12 weights to obtain our final model weights.

\section{Implementation Details}
\label{sec:details}

\begin{figure}[htb]
\centering
\includegraphics[width=85mm]{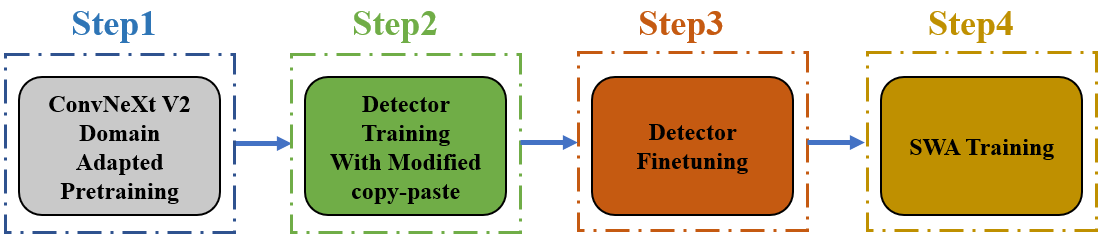}
\caption{Training process of only RGB modality input.}
\label{fig:RGB training steps}
\end{figure}

\textbf{Single-modality input.} In our experiments, training process of our best result for single RGB modality contains four steps described by Fig.~\ref{fig:RGB training steps}. 1) We employ the FCMAE method to pretrain a ConvNeXt V2 model for 100 epochs, with Imagenet22k pretrained weights as initialization. We combine the the RGB parts of the training data and validation data as our pretraining dataset. The model is trained under AdamW optimizer with a base learning rate of $10^{-4}$ and decay to $10^{-6}$ during the whole process. The main data augmentation techniques including random resizing, random cropping, and random flipping, the image mask ratio is 0.6 and L2 loss is used for supervision. 2) After pretraining, we train and finetune our fine-grained instance segmetor for 45 epochs and 5 epochs. 3) During finetuning, simple data augmentation strategies that do not involve modified copy-paste are employed. 4) After completing all training and finetuning steps, another 12 epochs SWA cyclical training procedure are adopted.

\textbf{Muti-modality input.} Several experiments are conducted to explore the potential of multi-modal inputs consisting of RGB and SAR data. However, the results of the multi-modal approach are not as good as those obtained with the single RGB modality, 

\begin{table*}[]
\centering
\caption{The details of our ablation study, CMR=Cascade Mask Rcnn, SCP=Simple Copy-Paste, MCP=Modified Copy-Paste, db=DUal-Backbone, DAP=Domain Adapted Pretraining, CNV2=ConvNeXt V2, WSF= Weighted Segmentation Fusion.}
\begin{tabular}{ccc}%
\toprule 
Method & \makecell[c]{Development phase (mAP50)}  &  \makecell[c]{Test phase (mAP50)} \\
\bottomrule 
CMR + db-swin-base & 0.443 & / \\
*\makecell[c]{CMR + db-swin-base \textbf{+ SAR-input}} & 0.371 & / \\
CMR + db-swin-base\textbf{+ seesaw} & 0.461 & / \\
\makecell[c]{CMR + db-swin-base + seesaw \textbf{+ SCP}} & 0.469 & / \\
\makecell[c]{CMR \textbf{+ db-CNV2-base-DAP}+ seesaw + SCP} & 0.476 & / \\
\makecell[c]{CMR + db-CNV2-base-DAP + seesaw + SCP \textbf{+ SWA}} & 0.485 & / \\
\makecell[c]{CMR + db-CNV2-base-DAP + seesaw \textbf{+ MCP} + SWA} & 0.496 & 0.426 \\
\makecell[c]{CMR + db-CNV2-base-DAP + seesaw \textbf{+ MCP+finetune} + SWA} & / & 0.441 \\
\makecell[c]{CMR + \textbf{db-CNV2-large-DAP} + seesaw + MCP+finetune + SWA} & / & 0.448 \\
\toprule 
\textbf{WSF based model ensemble} & 0.5510 & 0.5060 \\
\bottomrule 
\end{tabular}
\label{tab:ablation study}
\end{table*}

\section{Ablation study}
\label{sec:Experiments}

Table \ref{tab:ablation study} shows the ablation study. Our best result $mAP_{50}$ 50.6\% is acquired by fusing several strong detectors trained under different hyperparameters and backbones with WSF. From our experiments, we find that SAR data does not enhance model performance as shown in Table \ref{tab:ablation study}*, accuracy drops significantly compare to single RGB modality input.

\section{Conclusion}
\label{sec:Conclusion}
After comprehensive experiments, we build a robust model for building roof instance segmentation with domain adapted pretraining and dual backbone by utilizing optical satellite images as input. The misalignment and heterogeneity between optical and SAR data bring difficulties for multimodual data fusion and performance improve. More exploration is needed to investigate whether SAR image data can actually contribute to this scenario and how to make full use of them in the future.

\section{ACKNOWLEDGEMENT}
\label{sec:ACKNOWLEDGEMENT}
The authors would like to thank the IEEE GRSS Image Analysis and Data Fusion Technical Committee, Aerospace Information Research Institute, Chinese Academy of Sciences, Universität der Bundeswehr München, and GEOVIS Earth Technology Co., Ltd. for organizing the Data Fusion Contest. This work was supported in part by the Special Funds for Creative Research (No. 2022C61540).

\bibliographystyle{IEEEbib}
\bibliography{refs}

\end{document}